\documentclass[nohyperref]{article}

\usepackage{microtype}
\usepackage{graphicx}
\usepackage{subfigure}
\usepackage{booktabs} 

\usepackage{hyperref}

\usepackage[accepted]{template/icml2023}

\usepackage{template/algorithm}
\usepackage{template/algorithmic}

\usepackage{amsmath}
\usepackage{amssymb}
\usepackage{mathtools}
\usepackage{amsthm}

\usepackage[capitalize,noabbrev]{cleveref}

\usepackage{multirow}

\usepackage{arydshln}

\theoremstyle{plain}

\theoremstyle{definition}

\theoremstyle{remark}

\usepackage[textsize=tiny]{todonotes}

\icmltitlerunning{Disentangled representations for genomic discovery}

\begin{document}

\twocolumn[
\icmltitle{Evaluating unsupervised disentangled representation learning for\\ genomic discovery and disease risk prediction}



\begin{icmlauthorlist}
\icmlauthor{Taedong Yun}{1}
\end{icmlauthorlist}

\icmlaffiliation{1}{Google Research, Cambridge, MA 02142, USA}

\icmlcorrespondingauthor{Taedong Yun}{tedyun@google.com}

\icmlkeywords{Machine Learning, ICML}

\vskip 0.3in
]



\printAffiliationsAndNotice{}  

\newcommand{\ZZ}{{\mathbb{Z}}}
\newcommand{\KL}{D_{\text{KL}}}
\newcommand{\zz}{{\mathbf{z}}}
\newcommand{\xx}{{\mathbf{x}}}
\newcommand{\bvae}{$\beta$-VAE}
\newcommand\ty[1]{\textcolor[rgb]{0.09, 0.35, 0.74}{[Comment: #1]}}  %

\begin{abstract}
High-dimensional clinical data have become invaluable resources for genetic studies, due to their accessibility in biobank-scale datasets and the development of high performance modeling techniques especially using deep learning. Recent work has shown that low dimensional embeddings of these clinical data learned by variational autoencoders (VAE) can be used for genome-wide association studies and polygenic risk prediction. In this work, we consider multiple unsupervised learning methods for learning disentangled representations, namely autoencoders, VAE, $\beta$-VAE, and FactorVAE, in the context of genetic association studies. Using spirograms from UK Biobank as a running example, we observed improvements in the number of genome-wide significant loci, heritability, and performance of polygenic risk scores for asthma and chronic obstructive pulmonary disease by using FactorVAE or $\beta$-VAE, compared to standard VAE or non-variational autoencoders. FactorVAEs performed effectively across multiple values of the regularization hyperparameter, while $\beta$-VAEs were much more sensitive to the hyperparameter values.
\end{abstract}

\section{Introduction}
\label{sec:intro}


\begin{figure}[ht]
\vskip 0.2in
\begin{center}
\centerline{\includegraphics[width=\columnwidth]{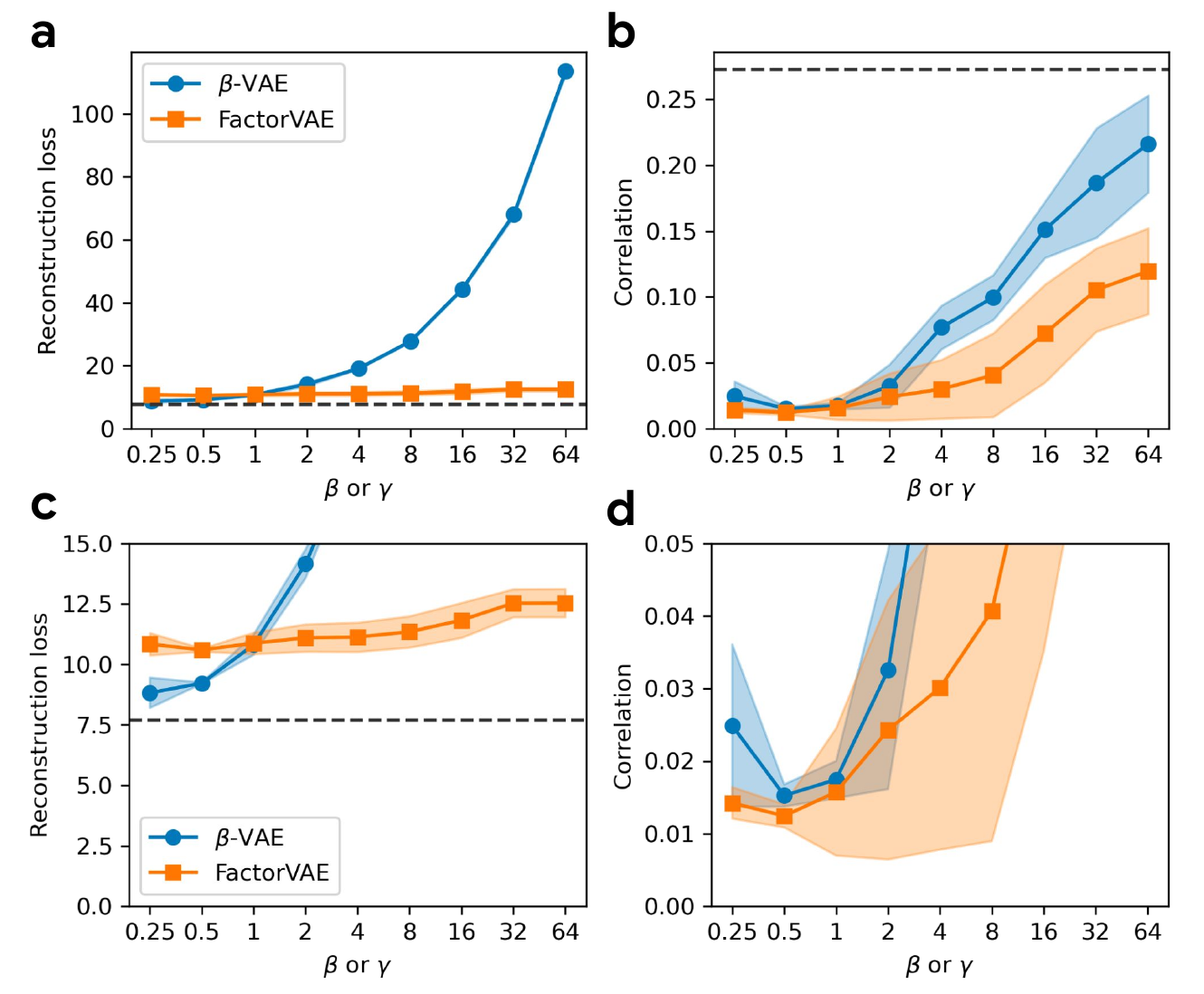}}
\caption{Reconstruction error and correlation between coordinates, as measured by the mean-squared error and the mean absolute values of correlations between all pairs of coordinates. We vary $\beta$ or $\gamma$ parameters in $\beta$-VAE and FactorVAE, which controls the strength of ``regularization''. Each setup is trained with 10 different random seeds. The data points are the mean of the 10 trials and the shaded regions are 95\% confidence intervals. The horizontal dashed line is from a non-variational autoencoder. \textbf{(a)} Reconstruction error in the validation set. \textbf{(b)} Mean of absolute values of correlations between all pairs of coordinates. \textbf{(c)} Same as (a), zoomed in. \textbf{(d)} Same as (b), zoomed in.}
\label{fig:reconstruction}
\end{center}
\vskip -0.2in
\end{figure}

\begin{figure*}[ht]
\vskip 0.2in
\begin{center}
\centerline{\includegraphics[width=0.9\textwidth]{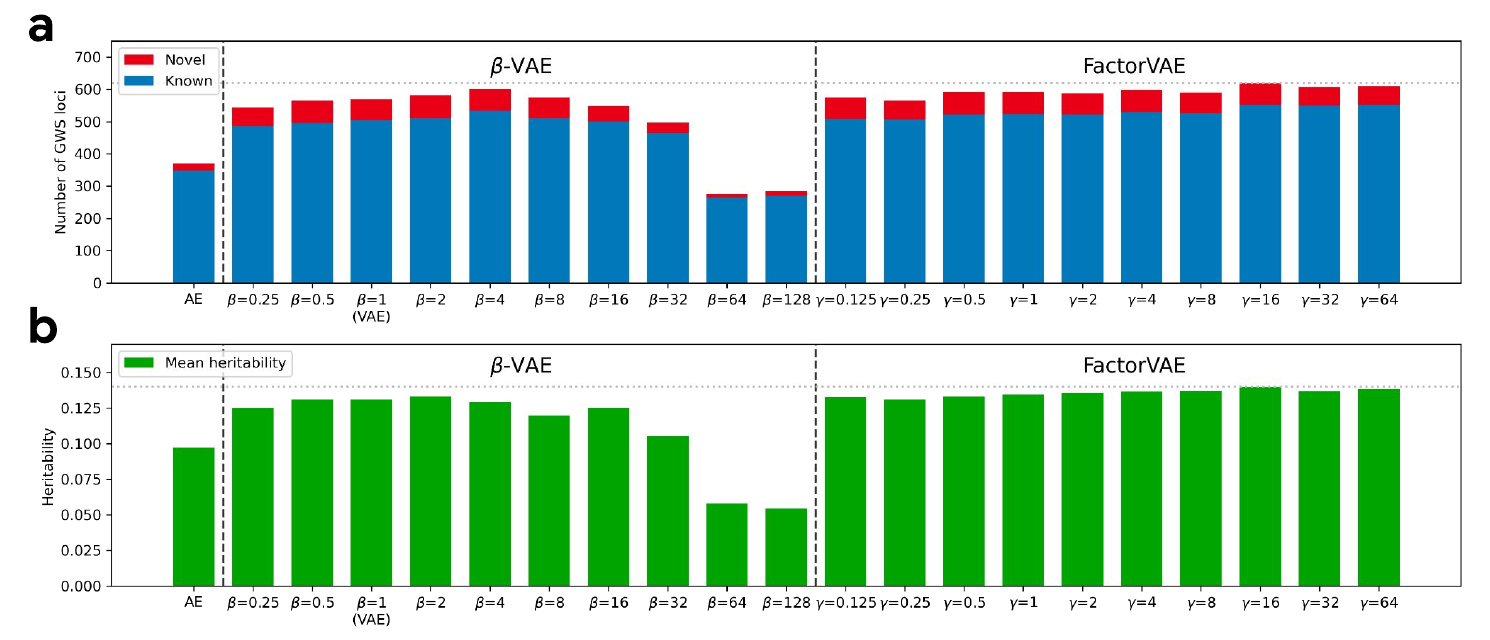}}
\caption{Genome-wide significant loci and heritability from GWAS on unsupervised representations. \textbf{(a)} Number of genome-wide significant loci. ``Known'' and ``novel'' is with respect to all loci from \cite{Shrine2023-tt} and GWAS Catalog lung function search. \textbf{(b)} Mean heritability ($h^2_g$) of all coordinates of the learned representation estimated by LD Score regression. Horizontal dotted line indicates the highest value. AE = (non-variational) autoencoder.}
\label{fig:gwas_loci_h2g}
\end{center}
\vskip -0.2in
\end{figure*}

\begin{figure*}[h]
\vskip 0.2in
\begin{center}
\centerline{\includegraphics[width=0.9\textwidth]{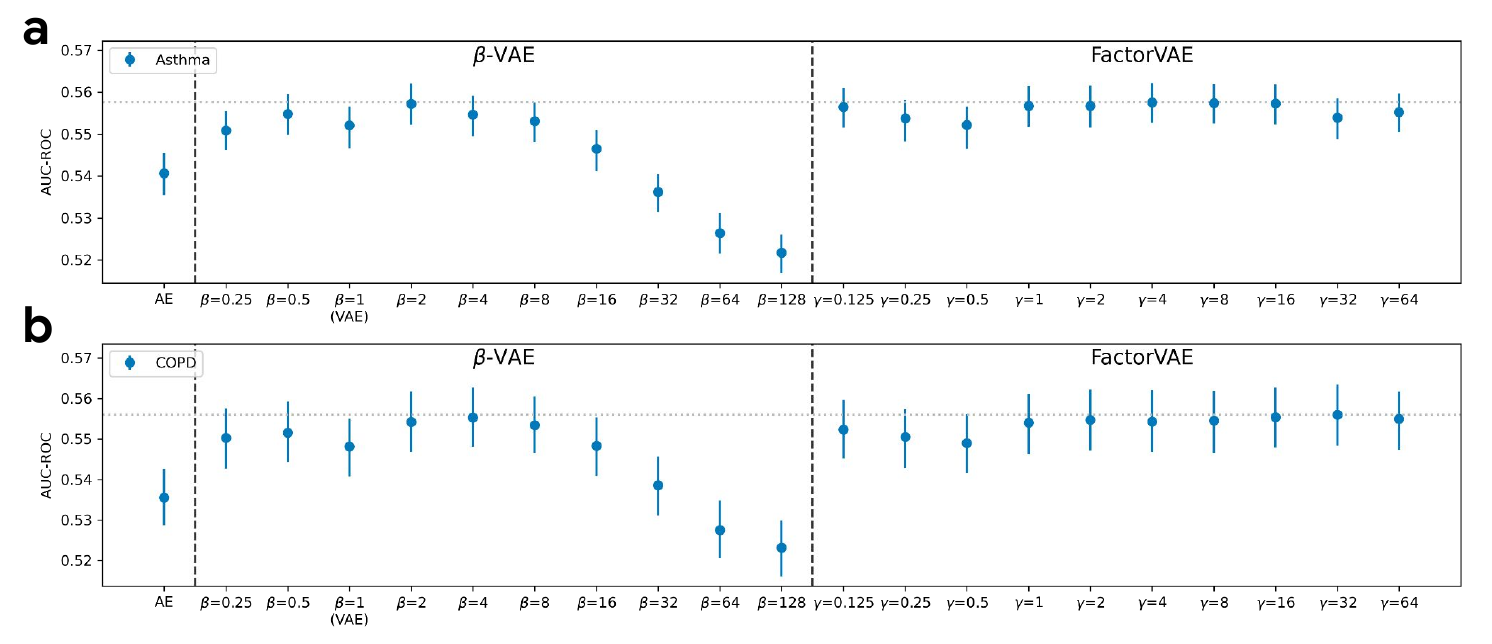}}
\caption{Polygenic risk score performance for asthma and COPD. We vary $\beta$ or $\gamma$ parameters in $\beta$-VAE and FactorVAE, which controls the strength of ``regularization''. Horizontal dotted line indicates the highest value. AE = (non-variational) autoencoder. \textbf{(a)} PRS performance in terms of AUC-ROC for asthma classification. \textbf{(b)} PRS performance in terms of AUC-ROC for COPD classification.}
\label{fig:prs}
\end{center}
\vskip -0.2in
\end{figure*}

\begin{table*}[h]
\caption{Estimated heritability and intercept from LD Score regression. The second column represents the $\beta$ values for \bvae{} and the $\gamma$ values for FactorVAE. The min, max, and mean is taken across five learned latent coordinates from each model. $h^2_g$ = heritability; $b$ = intercept.}
\label{tab:ldsc}
\vskip 0.15in
\begin{center}
\begin{small}
\begin{sc}
\begin{tabular}{cc|rrr|rrr}
\toprule
Model &    $\beta$ or $\gamma$ &  $\text{min}(h^2_g)$ &  $\text{max}(h^2_g)$ &  $\text{mean}(h^2_g)$ &  $\text{min}(b)$ &  $\text{max}(b)$ &  $\text{mean}(b)$ \\
\midrule
        AE &   - &   0.0531 &   0.1598 &    0.0971 &         1.0005 &         1.0382 &          1.0143 \\
\hdashline[1pt/3pt]

   \multirow{10}{*}{\bvae{}} &         0.25 &   0.0390 &   0.2239 &    0.1251 &         1.0053 &         1.0444 &          1.0275 \\
   &         0.5 &   0.0410 &   0.2463 &    0.1311 &         1.0046 &         1.0466 &          1.0300 \\
   &         1 &   0.0417 &   0.2515 &    0.1310 &         1.0051 &         1.0447 &          1.0283 \\
   &         2 &   0.0411 &   0.2628 &    0.1333 &         1.0026 &         1.0549 &          1.0303 \\
   &         4 &   0.0387 &   0.2582 &    0.1292 &         1.0031 &         1.0541 &          1.0262 \\
   &         8 &   0.0113 &   0.2536 &    0.1200 &         1.0013 &         1.0557 &          1.0277 \\
   &        16 &   0.0388 &   0.2428 &    0.1252 &         0.9874 &         1.0549 &          1.0156 \\
   &        32 &   0.0347 &   0.1931 &    0.1052 &         1.0022 &         1.0548 &          1.0263 \\
   &        64 &   0.0159 &   0.1866 &    0.0580 &         0.9966 &         1.0446 &          1.0146 \\
   &       128 &   0.0165 &   0.1865 &    0.0545 &         0.9979 &         1.0449 &          1.0101 \\
 \hdashline[1pt/3pt]

 \multirow{10}{*}{FactorVAE} &         0.125 &   0.0409 &   0.2507 &    0.1325 &         1.0048 &         1.0461 &          1.0300 \\
 &         0.25 &   0.0421 &   0.2526 &    0.1310 &         1.0023 &         1.0476 &          1.0291 \\
 &         0.5 &   0.0414 &   0.2583 &    0.1331 &         1.0037 &         1.0543 &          1.0306 \\
 &         1 &   0.0418 &   0.2597 &    0.1346 &         1.0026 &         1.0499 &          1.0306 \\
 &         2 &   0.0408 &   0.2589 &    0.1356 &         1.0042 &         1.0532 &          1.0313 \\
 &         4 &   0.0397 &   0.2547 &    0.1367 &         1.0055 &         1.0489 &          1.0308 \\
 &         8 &   0.0407 &   0.2661 &    0.1372 &         1.0019 &         1.0583 &          1.0331 \\
 &        16 &   0.0394 &   0.2657 &    0.1401 &         1.0034 &         1.0651 &          1.0331 \\
 &        32 &   0.0382 &   0.2638 &    0.1370 &         1.0040 &         1.0614 &          1.0328 \\
 &        64 &   0.0389 &   0.2647 &    0.1386 &         1.0041 &         1.0611 &          1.0349 \\
\bottomrule
\end{tabular}
\end{sc}
\end{small}
\end{center}
\vskip -0.1in
\end{table*}


Large-scale biobank projects such as UK Biobank with deep phenotyping and genotyping data enabled new frontiers in the study of human genetics \cite{Bycroft2018-pz}. High performance analysis methods using deep learning are well-suited for utilizing the high-dimensional clinical data (HDCD) (e.g. time series, images, videos) available in these datasets \cite{Bai2020-aw,Alipanahi2021-xa,Aung2022-ko,Pirruccello2022-ph,Cosentino2023-bx}.

A recent study \cite{Yun2023} demonstrated the potential for using HDCD in the context of genome-wide association studies (GWAS) to discover novel genetic insights about biological function, without any disease or trait labels. Using deep unsupervised representation learning techniques, specifically variational autoencoders (VAE) \cite{Kingma2013-vae}, the authors demonstrated new associations between genotypes and lung function captured by spirograms, a graphical representation of widely-used clinical pulmonary function test results. VAEs generate latent representations whose coordinates are relatively \emph{disentangled}, in which separable biological function can be better captured.

In this work, we extend this line of investigation by considering several modifications of VAE introduced to further increase the effect of disentanglement, namely \bvae{} \cite{Higgins2017-fi} and FactorVAE \cite{kim2018-pmlr}, in addition to standard VAE and (non-variational) autoencoders. We perform comprehensive evaluation of these methods in terms of reconstruction performance, correlation between coordinates (as a measure of disentanglement), and their performance in the context of GWAS for genomic discovery and polygenic risk prediction.

\section{Background}
\label{sec:background}

In representation learning of high-dimensional data, representations with ``disentangled'' coordinates are generally preferred so that independent factors of variations in the data can be separately captured by each coordinate \cite{Bengio2013-rl}. One of the most widely used methods for unsupervised learning of disentangled representation is VAE, in which the Kullback–Leibler (KL) divergence term $\KL(q(\zz | \xx)\parallel p(\zz))$ in the loss function implicitly encourages the learned approximate posterior distribution $q(\zz | \xx)$ to have independent coordinates when used with a factorized Gaussian prior distribution for $p(\zz)$. Several extensions or modifications to the VAE have been proposed to further enhance the disentanglement effect \cite{Higgins2017-fi,Burgess2018-ak,kim2018-pmlr,kumar2018,chen2018}. See \cite{Locatello2019-vr} for empirical comparison of these methods.

In this work, we focus our attention to two of these methods: \bvae{} \cite{Higgins2017-fi} and FactorVAE \cite{kim2018-pmlr}. \bvae{} amplifies the KL divergence term in the VAE loss function by a factor of $\beta > 1$, strengthening the ``regularization'' effect. With $\beta=1$, \bvae{} reduces to the standard VAE. On the other hand, FactorVAE adds an additional loss term to explicitly penalize the \emph{total correlation} (TC; a measure of dependence for multiple random variables) \cite{Watanabe1960-vl} of the latent coordinates, where an additional discriminator is jointly trained to approximate the TC using the density ratio trick \cite{Nguyen2010-fi,Sugiyama2012-me}. FactorVAE also has a hyperparameter $\gamma$ that controls the strength of the TC loss term.

Meanwhile, a recent work \cite{Yun2023} explored utilizing the disentangled representation of high-dimensional clinical data (HDCD) for genetic studies, specifically in the context of genome-wide association studies (GWAS) and polygenic risk scores (PRS). In their REGLE (REpresentation learning for Genetic discovery on Low-dimensional Embeddings) framework, they used VAEs to learn unsupervised disentangled representations of HDCD, e.g. spirograms, and performed GWAS on each coordinate to discover novel lung function loci in addition to recovering known lung function loci all without any disease labels. Moreover, using a (small) number of samples with paired genetics and disease labels (but not HDCD), the PRSs of the latent embeddings can be combined into a PRS specific to the given lung disease \emph{post hoc}.

\section{Experiments}
\label{sec:exp}

We used spirograms in UK Biobank \cite{Sudlow2015-mr,Bycroft2018-pz}, following the preprocessing steps and the dataset splits (train, validation, and PRS evaluation sets) described in \cite{Yun2023,Cosentino2023-bx}.

We compared convolutional (non-variational) autoencoders, VAEs, \bvae{}s, and FactorVAEs with identical encoder and decoder architecture (\cref{appx:sec:model_architecture}). We varied the $\beta$ parameter in \bvae{} and the $\gamma$ parameter in FactorVAE such that $\beta \in \{2^{i} ~\vert~ -2 \leq i \leq 7,~ i\in \ZZ \}$ and $\gamma \in \{2^{i} ~\vert~ -3 \leq i \leq 6,~ i\in \ZZ \}$. Roughly speaking, the $\beta$ and $\gamma$ values control the extent to which the loss function in \bvae{} and FactorVAE penalizes the entangled coordinates, higher values implying more penalization. The latent dimension was fixed to 5 to match \cite{Yun2023} for direct comparison.

Two main metrics were computed to compare the models, first without using the genetic data: the reconstruction error and the pairwise correlations of latent coordinates, both in the validation set (\cref{fig:reconstruction}). The reconstruct error was defined by the mean-squared error between the reconstructed spirogram and the original input spirogram. We also computed ``mean absolute correlation'' in the validation set defined by $\frac{\sum_{i < j} \vert \text{Corr}(\zz{}_i, \zz{}_j) \vert}{n(n-1)/2}$ for the latent vector $[\zz_i]_{i=1}^n$. We trained each model 10 times with different random seeds to compute 95\% confidence intervals ($\pm1.96$ $\times$ standard error). As also observed in \cite{kim2018-pmlr}, FactorVAEs retained good reconstruction across different values of the regularization hyperparameter $\gamma$, while \bvae{}'s reconstruction error significantly increased when $\beta$ was high (\cref{fig:reconstruction}a, c). Notably, the pairwise correlation metric also increased with very high values of $\beta$ or $\gamma$ (\cref{fig:reconstruction}b, d), implying less disentangled coordinates. For all remaining genetic analysis, we chose the trained model with the median reconstruction error in the validation set (among identical models trained with different random seeds), as a ``typical'' example of the representation generated in a similar setup.

We performed GWAS on all 5 coordinates of the learned representations using linear mixed model association testing implemented by BOLT-LMM \cite{Loh2015-ij}. Using the stratified linkage disequilibrium score regression (LDSC) \cite{BulikSullivan2015}, we observed high heritability ($h^2_g$ up to 27\%) of learned coordinates, especially from FactorVAEs (\cref{fig:gwas_loci_h2g}b, \cref{tab:ldsc}). The intercept term from LDSC was close to 1, indicating minimal confounding (\cref{tab:ldsc}).

Similarly to \cite{Yun2023}, we compared our genome-wide significant (GWS) loci to the largest known lung function GWAS \cite{Shrine2023-tt} and all lung function-related loci found in the NHGRI-EBI GWAS Catalog \cite{Sollis2023-pd} (\cref{appx:sec:catalog_search}). The independent GWS loci (linkage disequilibrium $R^2 \leq 0.1$ and $P \leq 5 \times 10^{-8}$) were defined by merging GWAS hits within 250kb together. We found that the encodings from FactorVAE generally produced more ``known'' and ``novel'' GWS loci compared to a plain VAE (i.e. \bvae{} with $\beta=1$), while the number of GWS loci varied significantly for \bvae{} across different values of $\beta$ (\cref{fig:gwas_loci_h2g}a).

Finally, using the REGLE framework introduced in \cite{Yun2023}, we generated PRSs specifically for two common lung-related diseases, asthma and chronic obstructive pulmonary disease (COPD), as a linear combination of 5 PRSs on learned latent coordinates. The weights of each latent coordinate PRS were learned using asthma and COPD labels in UK Biobank. As discussed in \cite{Yun2023}, one can use a small number of labels (as few as hundreds) to learn the weights. Comparing PRS prediction performance in terms of AUC-ROC in a held-out PRS evaluation set, we observed that PRSs generated by FactorVAEs consistently outperform PRS generated by a plain VAE or a non-variational autoencoder (\cref{fig:prs}). \bvae{} also outperformed the plain VAE with some values of $\beta$, but the performance was highly dependent on which $\beta$ value was chosen.

\section{Discussion}
\label{sec:discussion}

In this work, we performed comprehensive evaluation of multiple VAE-based methods for generating unsupervised disentangled representation, in the context of genomic discovery and polygenic risk prediction using high-dimensional clinical data (HDCD). To the best of our knowledge, this is the first such evaluation.

We studied how the different loss functions and the regularization hyperparameters in two types of VAE extensions (\bvae{} and FactorVAE) affect the ability to capture genetic components of the biological function encoded in spirograms, in addition to the reconstruction quality and the measure of disentanglement between coordinates. FactorVAEs were able to control the reconstruction error better, showed consistent ability to capture highly heritable coordinates, and discovered more genome-wide significant loci and more of the known loci than a standard VAE. While \bvae{} can also perform better than a standard VAE for certain values of $\beta$, the performance was highly sensitive to the choice of $\beta$. We observed the same patterns in PRS performance evaluation of asthma and chronic obstructive pulmonary disease based on the REGLE framework in \cite{Yun2023}.

We believe that unsupervised learning of HDCD is a promising direction for genetic studies and that high-performance representation learning methods will enhance our understanding of human genetics.

\section*{Software and Data}

Code available in \texttt{regle} directory at github.com/Google-Health/genomics-research.

\section*{Acknowledgements}

This research has been conducted using the UK Biobank Resource under Application Number
65275. We thank Farhad Hormozdiari, Cory Y. McLean, and Zachary R. McCaw for helpful discussions and comments.

\bibliography{refs}
\bibliographystyle{template/icml2023}

\newpage
\appendix
\onecolumn
\section{Model architecture and training}
\label{appx:sec:model_architecture}

The following model architecture was used for all VAE, \bvae{}, and FactorVAE experiments. The input shape is (1000, 2) representing volume-time and flow-time spirograms. FactorVAE has an additional discriminator trained jointly with the encoder and the decoder. Same architecture was used for (non-variational) autoencoders, except they do not have the final Gaussian sampling step (unique to VAE) in the encoder part. All models were implemented in TensorFlow 2 and Keras, and trained with Adam optimizer for 100 epochs, with learning rate of 0.0001 and batch size of 32.

\subsection{Encoder}

\begin{table*}[h]
\vskip 0.15in
\begin{small}
\begin{tabular}{lll}
\toprule
 Layer & Activation & Note \\ 
\midrule
 Conv1D & ReLU & filter=8, kernel=10, padding=same \\
 MaxPooling1D & -- & size=2 \\
 Conv1D & ReLU & filter=16, kernel=10, padding=same \\
 MaxPooling1D & -- & size=2 \\
 Conv1D & ReLU & filter=32, kernel=10, padding=same \\
 MaxPooling1D & -- & size=2 \\
 Flatten & -- & -- \\
 Dense & ReLU & size=64 \\
 Dense & ReLU & size=64 \\
 Dense & ReLU & size=64 \\
 Dense and Sampling & -- & size=5 for mean and variance each, and then sample from Gaussian. \\
\bottomrule
\end{tabular}
\end{small}
\vskip -0.1in
\end{table*}

\subsection{Decoder}

\begin{table*}[h]
\vskip 0.15in
\begin{small}
\begin{tabular}{lll}
\toprule
 Layer & Activation & Note \\ 
\midrule
 Dense & ReLU & size=64 \\
 Dense & ReLU & size=64 \\
 Dense & ReLU & size=64 \\
 Dense & ReLU & size=4000 \\
 Reshape & -- & shape=(125, 32) \\
 UpSampling1D & -- & size=2 \\
 Conv1DTranspose & ReLU & filter=16, kernel=10, padding=same \\
 UpSampling1D & -- & size=2 \\
 Conv1DTranspose & ReLU & filter=8, kernel=10, padding=same \\
 UpSampling1D & -- & size=2 \\
 Conv1DTranspose & ReLU & filter=2, kernel=10, padding=same \\
\bottomrule
\end{tabular}
\end{small}
\vskip -0.1in
\end{table*}

\subsection{Discriminator for FactorVAE}

\begin{table*}[h]
\vskip 0.15in
\begin{small}
\begin{tabular}{lll}
\toprule
 Layer & Activation & Note \\ 
\midrule
 Dense & LeakyReLU & size=1000 \\
 Dense & LeakyReLU & size=1000 \\
 Dense & LeakyReLU & size=1000 \\
 Dense & LeakyReLU & size=1000 \\
 Dense & LeakyReLU & size=1000 \\
 Dense & LeakyReLU & size=1000 \\
 Dense & -- & size=2 \\
 Softmax & -- & -- \\
\bottomrule
\end{tabular}
\end{small}
\vskip -0.1in
\end{table*}

\section{GWAS Catalog lung function search}
\label{appx:sec:catalog_search}

We used the following case-insensitive keywords to search for previously known lung-related GWAS loci in GWAS Catalog version \texttt{v1.0.2-associations\_e106\_r2022-07-09}: ``asthma'', ``chronic obstructive pulmonary disease'', ``copd'', ``expiratory flow'', ``fev1'', ``forced expiratory'', ``forced vital capacity'', ``lung function''.


\end{document}